# Self-aligned Spatial Feature Extraction Network for UAV Vehicle Re-identification

Aihuan Yao, Jiahao Qi, and Ping Zhong*, *Senior Member, IEEE*

*Abstract*—Compared with existing vehicle re-identification (ReID) tasks conducted with datasets collected by fixed surveillance cameras, vehicle ReID for unmanned aerial vehicle (UAV) is still under-explored and could be more challenging. Vehicles with the same color and type show extremely similar appearance from the UAV's perspective so that mining fine-grained characteristics becomes necessary. Recent works tend to extract distinguishing information by regional features and component features. The former requires input images to be aligned and the latter entails detailed annotations, both of which are difficult to meet in UAV application. In order to extract efficient fine-grained features and avoid tedious annotating work, this letter develops an unsupervised self-aligned network consisting of three branches. The network introduced a self-alignment module to convert the input images with variable orientations to a uniform orientation, which is implemented under the constraint of triple loss function designed with spatial features. On this basis, spatial features, obtained by vertical and horizontal segmentation methods, and global features are integrated to improve the representation ability in embedded space. Extensive experiments are conducted on UAV-VeID dataset, and our method achieves the best performance compared with recent ReID works.

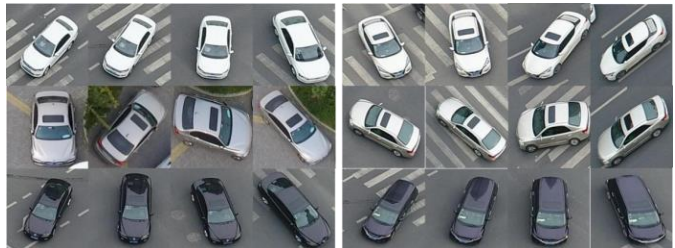

Fig. 1. Samples of UAV-VeID dataset. The first four images and the last four images come from different vehicles in each row.

## I. INTRODUCTION

BENEFITING from the rapid development of micro-electronics, navigation and communication technologies, UAV technology has made great progress, and are widely used in both military and civil fields, including air reconnaissance, patrol security and urban planning. The high-quality optical images obtained by UAV are used for computational analysis to realize object detection and scene classification [1]-[3], which is of great significance in the field of intelligent security. Vehicle re-identification (ReID), aiming at locating query vehicles from gallery set accurately, works as a key technology for social public security and smart city construction.

Existing ReID works mainly focus on datasets collected by surveillance cameras, which possess fixed locations and limited viewpoints. It is the key to mine distinguishing characteristics for vehicle ReID. Many fine-grained feature extraction methods have emerged consequently, which can be generally divided into three categories: attention mechanism-based [4], regional feature-based [5] and component feature-based [6][7]. Attention mechanism can help models highlight valid features and has been widely used for image processing. Considering the imbalance of discriminative features in different spatial locations and different channels, spatial and channel attention network (SCAN) was proposed [4]. What's more, as local region conveys more distinctive visual cues, Region-Aware deep Model (RAM) also extracts features from a series of regions [5], to encourage the deep model to learn more discriminative features. Besides, some of works utilize detailed annotation to find out key-parts positions to highlight effective component characteristics [6], including window, light and brand.

Vehicle ReID for UAV remote sensing images is still in the exploration stage in terms of methods and datasets. Teng contributed a novel dataset called UAV-VeID as shown in Fig.1 [8], proposed a viewpoint adversarial training strategy and a multi-scale consensus loss to promote the robustness and discriminative power of learned deep features simultaneously. From Fig. 1, we can see that vehicle images sharing the same color and type demonstrate excessive similarity but variable vehicle orientation, which raises a high demand for efficient feature extraction method. Whereas regional feature-based methods and component feature-based methods have the limitations of high dependence on input alignment and detailed annotations respectively, which are not suitable for the ReID in UAV application.

This work was supported in part by the Natural Science Foundation of China under Grant 61971428. (*Corresponding author: Ping Zhong*.)
Aihuan Yao is with the National Key Laboratory of Science and Technology on Automatic Target Recognition, National University of Defense Technology, Changsha 410073, China (e-mail: yaoah@nudt.edu.cn).
Jiahao Qi is with the National Key Laboratory of Science and Technology on Automatic Target Recognition, National University of Defense Technology, Changsha 410073, China (e-mail: qijiahao1996@163.com).
Ping Zhong is with the National Key Laboratory of Science and Technology on Automatic Target Recognition, National University of Defense Technology, Changsha 410073, China (e-mail: zhongping@nudt.edu.cn).

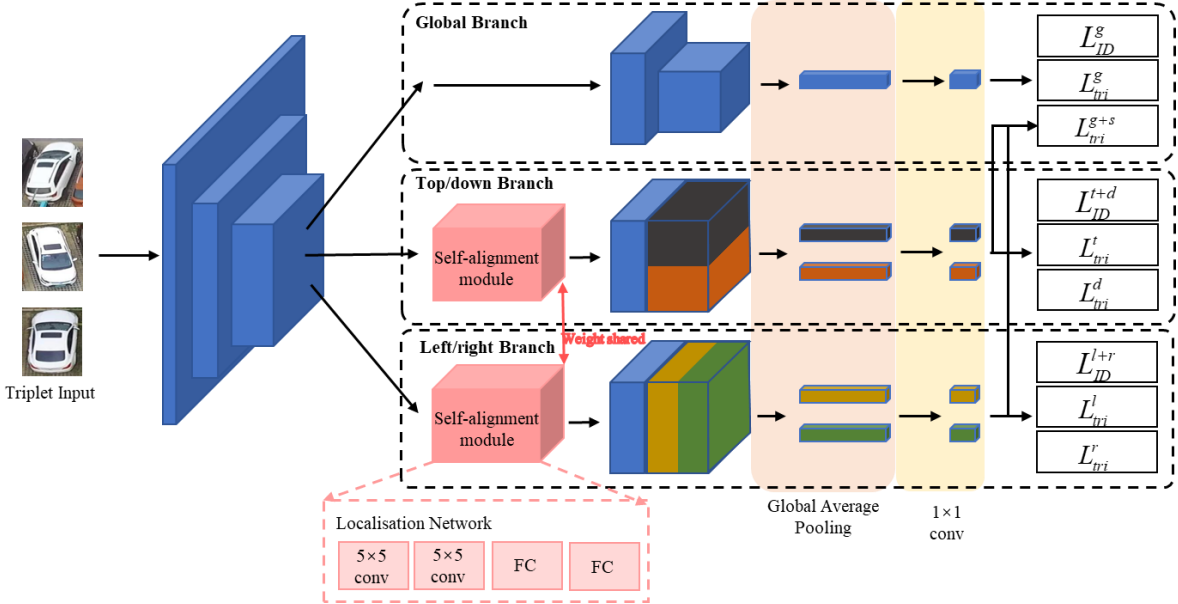

Fig. 2. The main structure of proposed method.

According to the above analysis, this letter proposes a self-aligned spatial feature extraction network (SANet), which can make deep network focus on feature extraction among a certain region and improve its representation ability. The primary contributions of our research are summarized as: on the one hand, we introduce a network with three branches to integrate global and spatial features, for sake of extracting fine-grained features. On the other hand, we introduce a self-alignment module and design loss function to align input images, which makes it feasible to segment images according to a certain standard without viewpoint annotations.

## II. METHODOLOGY

As mentioned, this letter aims to extract effective fine-grained features without relying on annotations. Therefore, we design the whole structure and the corresponding loss function to meet it.

### A. Overall Framework

Our deep network architecture consists of three branches: global branch, upper/lower branch, and left/right branch. Obviously, the former is used to extract global features, and the latter two are used to extract spatial features in different directions. Input images have different deep features after three parallel network branches, and we concatenate them to obtain the final representation vector. The overall architecture is illustrated in Fig. 2.

**Global Branch:** It is worth mentioning that the first four blocks in Resnet-50 are selected as shared shallow networks. In addition, the global branch is composed of block_5 in Resnet-50, pooling layer and 1×1 convolution layer, which is used to reduce the dimension of embedded features.

**Top/down Branch:** After shared shallow networks, there is a self-alignment module in this branch to convert the input images to a uniform orientation. What's more, the step size of spatial feature extraction is set to 1 for the sake of improving the resolution of feature maps. Then, we get the top and down spatial features in the way of horizontal segmentation.

**Left/right Branch:** It possesses the same architecture as another spatial branch and shares the parameters of the self-alignment module with it. In particular, we use vertical segmentation to obtain the left and right spatial features.

### B. Self-alignment Module

Considering the gigantic diversity in vehicle orientations, it is fundamental to align input images for extracting spatial features. This part is introduced to implement spatial transform [9], which can convert the input to a uniform orientation without relying on the viewpoint annotations. Consequently, it is feasible to divide vehicle areas in a fixed manner in the subsequent network.

The self-alignment module takes the output feature maps of shallow shared neural networks as input, and outputs the self-aligned feature maps. The calculation process could be depicted as follows. Firstly, a customized localisation network utilizes the input feature maps $U$ to generate the adaptive transform parameters $\theta$: $\theta = f_{loc}(U; W_{loc})$, where $f_{loc}$ is the localisation network function and $W_{loc}$ is corresponding parameters. In this letter, the localisation network consists of two convolutional layers and two fully connected layers as shown in Fig. 2. Then the grid generator creates a regular sampling grid $G = \{G_i\}$ according to $\theta$, where $G_i = (x_i^t, y_i^t)$ denotes the pixels on the output feature maps $V$. Therefore, the spatial transform between feature maps can be formulated as:

$$\begin{pmatrix} x_i^s \\ y_i^s \end{pmatrix} = T_\theta(G_i) = \begin{bmatrix} \theta_{11} & \theta_{12} & \theta_{13} \\ \theta_{21} & \theta_{22} & \theta_{23} \end{bmatrix} \begin{pmatrix} x_i^t \\ y_i^t \\ 1 \end{pmatrix} \quad (1)$$

where $(x_i^s, y_i^s)$ and $(x_i^t, y_i^t)$ are the coordinates in the input feature maps $U$ and output feature maps $V$ respectively. The main demand of spatial transform from UAV's perspective is to achieve self-alignment by rotating vehicle targets, therefore affine transformation is selected in this letter to meet this requirement. Finally, *V* could be obtained by means of differential image sampling. It is obvious that annotations are not necessary during aforementioned computational process. The aligned feature maps are more suitable for UAV vehicle ReID tasks when extracting spatial features.

*C. Objective Function*

Firstly, we regard the ReID task as a multi-classification problem, and carry out classification loss constraints on the features extracted from the three branches respectively:

$$\begin{aligned} L_{ID}^g &= -\frac{1}{N}\sum_{i=1}^{N}\sum_{c=1}^{M} y_{ic} \log(p_{ic}) \\ &= -\frac{1}{N}\sum_{i=1}^{N}\sum_{c=1}^{C} y_{ic} \log \frac{e^{W_{y_{ic}}^T f_i^g + b_{y_{ic}}}}{\sum_{k=1}^{C} e^{W_k^T f_i^g + b_k}} \end{aligned} \quad (2)$$

It shows the classification loss for the global branch, where *N* and *C* represent the number of training samples in mini-batch and the total number of vehicle ids, respectively. Besides, $y_i^c$ is a symbolic function, which equals 1 when the real category of sample *i* is equal to *c*, and 0 otherwise, $p_i^c$ denotes the probability that the observed sample *i* belongs to category *c*. Similarly, the classification loss for other branches, $L_{ID}^{t+d}$ and $L_{ID}^{l+r}$, can be calculated by concatenating two spatial features respectively.

Secondly, it is impossible to achieve self-alignment of input images only by triplet loss constraint on global features, which can be proved by related ReID methods who employ both triplet loss and classification loss. Therefore, in addition to the metric loss $L_{tri}^g$ for global features, this letter also sets corresponding metric loss for top, down, left and right spatial features respectively, which can be formulated as:

$$L_{tri}^g = max\left(\left\|f_a^g - f_p^g\right\|_2 - \left\|f_a^g - f_n^g\right\|_2 + m, 0\right) \quad (3)$$

$$L_{tri}^t = max\left(\left\|f_a^t - f_p^t\right\|_2 - \left\|f_a^t - f_n^t\right\|_2 + m, 0\right) \quad (4)$$

The other three losses $L_{tri}^d$, $L_{tri}^l$, $L_{tri}^r$ and $L_{tri}^{g+s}$ have a similar form, which encourages the distance of positive pairs to be smaller than negative sample pairs by a threshold *m*. Using triplet loss functions to make the segmented spatial features gather in the embedded space, the self-aligned module can generate the adaptive parameters according to images with variable orientations, so as to guarantee the consistency of spatial features. In other words, the triplet loss functions designed for spatial features replace the viewpoint annotations to some extent.

*D. Implementation Details*

As mentioned above, pre-trained Resnet-50 is selected as backbone network. Input images are resized to $256 \times 256$, and two methods of random erasing and color jitter are adopted. The margin of the triplet loss is set to 0.3, and the batch size is set to $32 = 8 \times 4$, indicating that 4 images of 8 vehicle ids are randomly selected in each batch. Adam optimizer with an initial learning rate of 0.0005 is adopted, which is adjusted according to cosine annealing strategy dynamically. Since the convergence result of self-aligned module in the network has a great impact on network performance, the learning rate of the spatial transform module of spatial feature extraction branch is set as $0.0005 \times 0.05$ to avoid the failure of convergence due to excessive initial learning rate. In the process of training, considering that Resnet-50 loads pre-training parameters, we only optimize the parameters of spatial transform module, $1 \times 1$ dimensional reduction convolution layer and the full connection layer during the first 10 epochs. Experiments are implemented based on NVIDIA 2080Ti GPU with 12GB RAM.

III. EXPERIMENTS

In this section, abundant experiments have been carried out on UAV vehicles dataset to testify the effectiveness of our method. Firstly, we briefly introduce necessary information about UAV-VeID [8] dataset and primary evaluation metric employed in this letter. Secondly, we analyze carefully the influence of the spatial features by conducting empirical studies, and the results of spatial transformation by visualization as well. At last, comparison experiments are conducted on UAV-VeID dataset and compared with recent works.

*A. Datasets and Evaluation Metrics*

**Datasets.** We firstly review necessary information about UAV vehicle dataset used in this letter, which is collected by UAV-mounted cameras.

UAV-VeID is constructed from video sequences captured by drones in different locations, backgrounds and lighting conditions, such as highway intersections, road crossings, parking lots, etc. The flight altitude of UAV is between 15 and 60 meters, and the vertical angle of the camera is between 40 and 80 degrees, resulting in the variable sizes and perspectives of vehicle targets. UAV-VeID dataset contains 41,917 images of 4,601 vehicles, of which the corresponding images of 1,797 vehicles, 596 vehicles and 2208 vehicles are used as training set, verification set and test set, respectively.

**Evaluation Metrics.** During the test, the Euclidean distance between the embedded features of query set and gallery set is firstly calculated and sorted in ascending order, and then the distance metric matrix is returned. Because there is only one ground truth match for a given query in UAV-VeID test set, we use only CMC-k to evaluate method performances, showing the probability of correct matching in the top-k ranked retrieved results. CMC-k calculates the mean top-k accuracy of all queries, which can be calculated as:

$$Acc_k = \begin{cases} 1 & if\ top-k\ gallery\ samples\ contain\ the\ query\ identity \\ 0 & otherwise \end{cases} \quad (5)$$

Since it only considers the first match in evaluation process, it is accurate for datasets where there is only one ground truth match for a given query, such as UAV-VeID dataset.

*B. Model Analysis*

In this subsection, we analyze the number of spatial feature blocks and then visualize the results of self-alignment.

The empirical study on spatial feature blocks of our SANet is implemented on UAV-VeID test set and the results are exhibited in Table I, where M denotes the number of blocks obtained during segmentation in the spatial branch of SANet, dim represents the dimension of integrating global and spatial features. It can be seen that there is little difference in ReID performance under the three kinds of spatial feature blocks, which can be interpreted as the high-dimensional features may contain more redundant information. Considering that in the process of vehicle ReID, CMC-1 evaluation metric has more practical application value, and the lower feature dimension is conducive to realizing vehicle ReID between larger databases. Therefore, we select SANet when M is 2 as the final network structure.

TABLE I
PERFORMANCE WITH DIFFERENT SPATIAL FEATURE BLOCKS ON UAV-VEID TEST SET. THE BEST VALUES ARE IN BOLD

| M | dim | CMC-1 | CMC-5 | CMC-10 |
|---|---|---|---|---|
| 2 | 2560 | **74.82** | 89.38 | 93.06 |
| 4 | 4608 | 73.52 | 89.58 | **93.16** |
| 6 | 6656 | 74.32 | 89.69 | 93.03 |

TABLE II
COMPARISON WITH RECENT WORKS ON UAV-VEID TEST SET. THE BEST VALUES ARE IN BOLD

| Method | CMC-1 | CMC-5 | CMC-10 |
|---|---|---|---|
| CN-Nets | 55.91 | 76.54 | 82.46 |
| RAM | 45.26 | 59.35 | 64.07 |
| SCAN | 40.49 | 53.74 | 60.55 |
| VSCR | 70.59 | 88.33 | 92.51 |
| AM+WTL | 69.11 | 87.23 | 91.64 |
| Baseline | 65.36 | 83.74 | 88.88 |
| SANet (ours) | **74.94** | **90.21** | **93.29** |

To testify whether the introduction of the self-alignment module achieved the expected effect, we randomly select several vehicle images of UAV-VeID test set and record the spatial transform parameters regressed by localisation network. Then the parameters are directly applied to the input image to verify the result of self-alignment, as illustrated in Fig. 3. As we can see, for input images with changeful views, the method proposed in this letter basically realizes unsupervised self-alignment, and can uniformly convert the input images to the "upper right" orientation. This result can provide guarantee for subsequent spatial feature extraction.

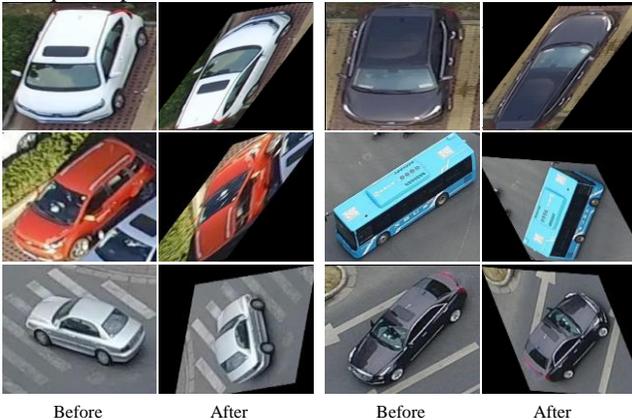

Before    After    Before    After

Fig. 3. Visualization of self-alignment results. Each row shows two groups of vehicle images before and after the spatial transformation.

### C. Comparison with Recent Works

As mentioned above, this letter uses CMC-k index to evaluate the performance of ReIDs algorithm on UAV-VeID test set, and compares it with several typical algorithms, as shown in table II. Among them, baseline model is trained by classification and triplet loss with ResNet-50 as backbone. Obviously, our method (SANet) achieves the best CMC-1 accuracy of 74.94%, outperforming other compared methods. It is interesting to observe that, CN-Nets [10], designed for fine-grained image retrieval with a coarse-to-fine framework, shows a better performance compared with RAM [5] and SCAN [4]. Among them, RAM obtains regional features through horizontal segmentation directly, which is not suitable for the situation where vehicle orientations cannot be determined from the perspective of UAV. SCAN takes advantage of attentional mechanism to extract regional features but performs poorly. This can be interpreted as algorithms designed for vehicle ReID tasks on traditional surveillance videos do not consider the viewpoint variety in UAV-VeID dataset. VSCR [8] and AM+WTL [11] utilize perspective annotations and color prior information respectively. The former uses adversarial learning to generate features that are robust to perspective changes, while the latter emphasizes the target and reduces the background through attention mask. However, both of them are not as effective as method proposed in this letter, which indicates that correct fine-grained feature extraction is more conducive to distinguishing similar vehicles. In general, SANet relies on no detail annotation and achieves the optimal performance of UAV vehicle ReID.

More intuitively, Fig. 4 shows the CMC curves of each algorithm on the UAV-VeID test set, which shows the probability of correct results in the top k images in the ReID results. Therefore, the closer the CMC curve is to the upper left corner, the stronger the performance of the corresponding algorithm is. The figure shows that the effect of SANet algorithm is generally superior to other comparison algorithms. Specifically, the hit rate of SANet Vehicle Association before rank-10 is higher than VSCR, and the performance of the two algorithms after rank-10 is similar.

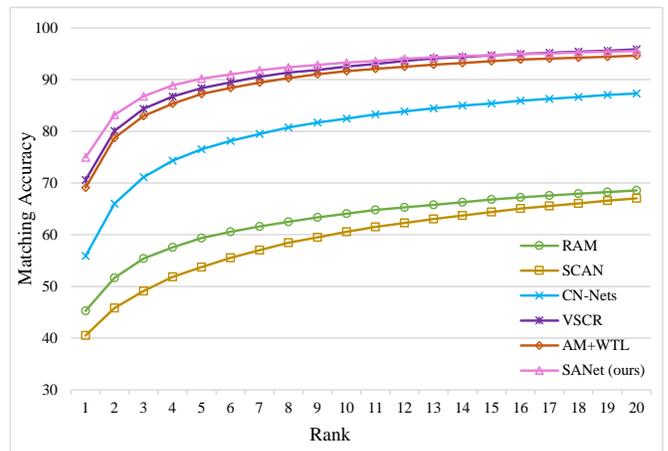

Fig. 4. CMC curves of compared methods on UAV-VeID.

Furthermore, we qualitatively show the feature embedding space obtained by different algorithms to demonstrate the validity of our methods. The more vehicle features with the same ID in the embedding space gather, the more distinctive and robust the features learned by this network are. In this letter, vehicle images of 15 IDs in the UAV-VeID test set are randomly selected. As shown in Fig. 5, t-SNE [12] dimension reduction method is adopted to reduce the obtained 2560-dimensional representation vector to 2-dimensional vector. It can be seen that algorithm proposed in this letter can make multiple images with the same ID more compact in the embedded space, which is more conducive to distinguish similar vehicle targets.

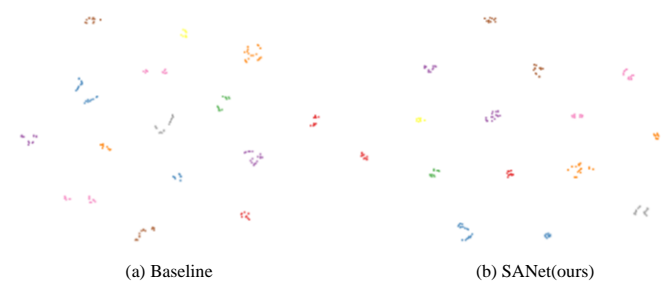

Fig. 5. Visualization with t-SNE for feature distribution of 15 vehicles in UAV-VeID dataset (best viewed in color).

At last, we demonstrate some vehicle ReID results on UAV-VeID in Fig. 6, where the baseline and the SANet are compared. As we can see, our proposed method has better capability to distinguish the match results from similar false positives in gallery set compared with our baseline model. It demonstrate that our model learns more efficient and robust vehicle features.

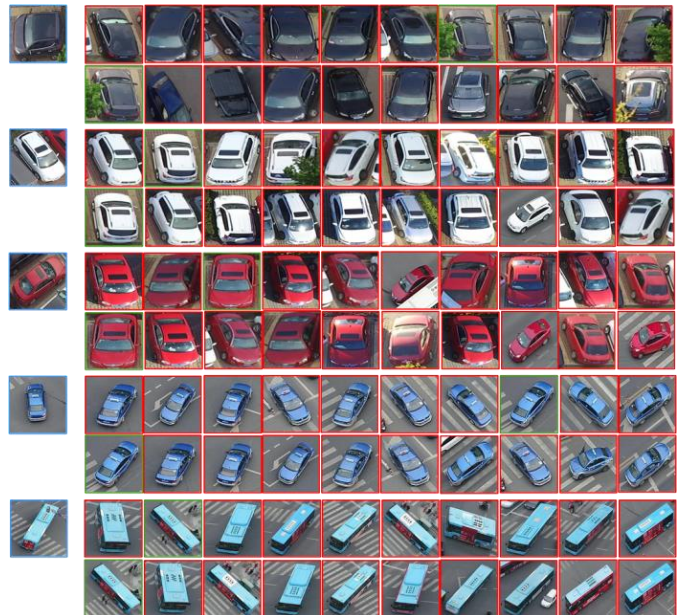

Fig. 6. Vehicles ReID results on UAV-VeID. The first and second row denote top-10 returned images retrieved by baseline model and our SANet model, respectively. Blue boxes represent queries, green and red boxes represent true positives and false positives.

## IV. CONCLUSION

In this letter, we analyzed the characteristics of vehicle images from the perspective of UAV firstly. Aiming at the variability of vehicle target orientation, we introduced self-alignment module and designed triplet loss with spatial features to realize the alignment of input images without annotations. On this basis, we formulated a vertical and horizontal segmentation method to extract spatial features, and integrate global features and spatial features simultaneously to improve the ability of representation in the embedded space. Extensive experiments verified that our method achieved the best performance on UAV-VeID dataset compared with recent works. In general, our model proposed in this letter does not rely on any additional annotation but achieves optimal performance in terms of evaluation metrics, which is more potential to be applied to real-world scenarios.

## V. REFERENCE

[1] G. Cheng, Y. Si, H. Hong, X. Yao and L. Guo, "Cross-Scale Feature Fusion for Object Detection in Optical Remote Sensing Images", *IEEE Geosci. Remote Sens. Lett.*, vol. 18, no. 3, pp. 431-435, 2021.

[2] S. Zhang, G. He G, H. Chen, "Scale adaptive proposal network for object detection in remote sensing images", *IEEE Geosci. Remote Sens. Lett.*, vol. 16, no. 6, pp. 864-868, 2019.

[3] A. E. Almeida and R. d. S. Torres, "Remote Sensing Image Classification Using Genetic-Programming-Based Time Series Similarity Functions", *IEEE Geosci. Remote Sens. Lett.*, vol. 14, no. 9, pp. 1499-1503, 2017.

[4] S. Teng, X. Liu, S. Zhang and Q. Huang, "SCAN: Spatial and channel attention network for vehicle re-identification", *Proc. Pacific Rim Conf. Multimedia*, pp. 350-361, 2018.

[5] X. Liu, S. Zhang, Q. Huang and W. Gao, "Ram: A region-aware deep model for vehicle re-identification", *Proc. ICME*, pp. 1-6, Jul. 2018.

[6] B. He, J. Li, Y. Zhao, and Y. Tian, "Part-regularized near-duplicate vehicle re-identification," in *Proc. IEEE/CVF Conf. Comput. Vis. Pattern Recognit. (CVPR)*, pp. 3997–4005, Jun. 2019.

[7] P. Khorramshahi, A. Kumar, N. Peri, S. S. Rambhatla, J.-C. Chen and R. Chellappa, "A dual-path model with adaptive attention for vehicle re-identification", *Proc. IEEE Int. Conf. Comput. Vis. (ICCV)*, Oct. 2019.

[8] S. Teng, S. Zhang, Q. Huang and N. Sebe, "Viewpoint and Scale Consistency Reinforcement for UAV Vehicle Re-Identification", *Int J Comput. Vis.*, 129, 719-735, 2021.

[9] M. Jaderberg, K. Simonyan, A. Zisserman, "Spatial transformer networks", *Adv. Neural. Inf. Process. Syst. (NIPS)*, 28, 2017-2025, 2015.

[10] H. Yao, S. Zhang, Y. Zhang, J. Li and Q. Tian, "One-shot fine-grained instance retrieval", *Proc. ACM Multimedia Conf. (MM)*, pp. 342-350, 2017.

[11] A. Yao, M. Huang, J. Qi and P. Zhong, "Attention Mask-Based Network With Simple Color Annotation for UAV Vehicle Re-Identification". *IEEE Geosci. Remote Sens. Lett.*, 2021.

[12] L Maaten, G. Hinton, "Visualizing Data using t-SNE", *J. Mach. Learn. Res.*, pp. 2579-2605, 2008.